\documentclass[conference]{IEEEtran}
\IEEEoverridecommandlockouts

\usepackage[mode=buildnew]{standalone} % standalone compilation
\usepackage{cite}                       % IEEE style citations
\usepackage{textcomp}                   % extra symbols
\usepackage{xcolor}                     % colors
\usepackage[inline]{enumitem}           % inline enumerations
\setlist[enumerate,1]{label=(\roman*)}  % roman numerals for first-level enumerate
\usepackage[caption=false]{subfig}
\usepackage{balance}
\usepackage{amsmath, amssymb, amsfonts, amsthm}
\usepackage{siunitx}                    % units
\usepackage{algorithmic}
\usepackage[ruled,vlined]{algorithm2e}

\usepackage{graphicx}
\usepackage{tikz}
\usetikzlibrary{
    arrows.meta,
    backgrounds,
    bending,
    calc,
    math,
    matrix,
    patterns,
    patterns.meta,
    positioning,
    quotes,
    shapes,
    shapes.geometric,
    shapes.multipart,
    angles
}
\tikzset{
    every picture/.style={/utils/exec={\footnotesize}},
    every node/.style={font=\footnotesize},
    arrow/.style={-Stealth},
    edge/.style={-Stealth,shorten >=1pt},
    biedge/.style={-,shorten >=1pt,shorten <=1pt},
    coupling/.style={
        matrix of nodes,
        column sep={12mm,between origins},
        row sep={12mm,between origins},
        nodes={circle, draw, minimum size=6mm}
    }
}

\usepackage{pgfplots}
\pgfplotsset{compat=1.18}
\usepgfplotslibrary{fillbetween, groupplots, units, statistics, colormaps}
\usepackage{pgfplotstable}
\pgfdeclarelayer{bg}   
\pgfsetlayers{bg,main}

\pgfplotsset{
    compat=1.18,
    my boxplot style/.style={
        boxplot,
        mark=x,
        solid
    },
}  
\pgfplotsset{ 
    /pgfplots/custom legend/.style={
        legend image code/.code={
            \draw[##1,only marks,mark=square*,mark options={solid}]
                plot coordinates {(0.3cm,0cm)};
        },
        column sep=0.1cm
    },
}

\usepackage{booktabs}
\newcolumntype{L}[1]{>{\raggedright\arraybackslash}p{#1}}
\newcolumntype{C}[1]{>{\centering\arraybackslash}p{#1}}
\newcolumntype{R}[1]{>{\raggedleft\arraybackslash}p{#1}}

\usepackage{hyperref}
\usepackage[nameinlink, capitalise]{cleveref}

\usepackage[nolist]{acronym}
\begin{acronym}
    \acro{CPM Lab}{Cyber-Physical Mobility Lab}
    \acro{CAV}{Connected and Automated Vehicle}
    \acro{OSN}{Outdoor Sensor Node}
    \acro{SSL}{Sensitive Surface Layer}
    \acro{RSU}{Roadside Unit}
    \acro{Ufil}{Unified Framework for Infrastructure-based Localization}
    \acro{ROS 2}{Robot Operating System 2}
    \acro{CAM}{Cooperative Awareness Messages}

    \acro{V2I}{Vehicle-to-Infrastructure}
    \acro{V2X}{Vehicle-to-Everything}
    \acro{FOV}{Field of View}
    \acro{RMSE}{Root Mean Square Error}

\end{acronym}

\usepackage{orcidlink}

\def\BibTeX{{\rm B\kern-.05em{\sc i\kern-.025em b}\kern-.08em
    T\kern-.1667em\lower.7ex\hbox{E}\kern-.125emX}}

\newcommand{\currentassociationmatrix}{{\pmb{A}_{k}}}

\newcommand{\currentcostmatrix}{{\pmb{C}_{k}}}
\newcommand{\costmatrixthreshold}{{c_{max}}}
\newcommand{\covariancematrix}{{\pmb{P}}}

\newcommand{\objectlist}{{\mathcal{O}}}
\newcommand{\currentobjectlist}{{\mathcal{O}_{k}}}

\newcommand{\object}{{O}}

\newcommand{\statevector}{{\pmb{x}}}
\newcommand{\stateinput}{{\pmb{u}}}
\newcommand{\estimatedstatevector}{{\hat{\statevector}}}
\newcommand{\statecovariancematrix}{{\covariancematrix}_{\estimatedstatevector}}
\newcommand{\statetransitionmatrix}{{\pmb{F}}}
\newcommand{\stateinputmatrix}{{\pmb{B}}}
\newcommand{\processnoisematrix}{{\pmb{Q}}}
\newcommand{\measurementnoisematrix}{{\pmb{R}}}
\newcommand{\measurementmatrix}{{\pmb{H}}}
\newcommand{\innovationmatrix}{{\pmb{S}}}

\newcommand{\positionvector}{{\pmb{p}}}
\newcommand{\positionx}{{p_x}}
\newcommand{\positiony}{{p_y}}

\newcommand{\velocityvector}{{\pmb{v}}}
\newcommand{\velocityx}{{v_x}}
\newcommand{\velocityy}{{v_y}}

\newcommand{\measurementvector}{{\pmb{z}}}
\newcommand{\innovationvector}{{\pmb{y}}}

\newcommand{\yaw}{{\theta}}
\newcommand{\yawrate}{{\omega}}

\newcommand{\dimensionvector}{{\pmb{d}}}

\newcommand{\length}{{w}}
\newcommand{\width}{{l}}
\newcommand{\height}{{h}}

\newcommand{\estimateddimensionvector}{{\hat{\dimensionvector}}}

\newcommand{\dimensionvariancematrix}{{\covariancematrix}_{\estimateddimensionvector}}

\newcommand{\probability}{{p}}
\newcommand{\existenceprobability}{{\probability_e}}
\newcommand{\classificationvector}{{\pmb{c}}}

\definecolor{color1}{RGB}{0, 84, 159}   
\definecolor{color2}{RGB}{246, 168, 0}  
\definecolor{color3}{RGB}{0, 97, 101}   
\definecolor{color4}{RGB}{227, 0, 102}  
\definecolor{color5}{RGB}{0, 152, 161}  
\definecolor{color6}{RGB}{204, 7, 30}   
\definecolor{color7}{RGB}{87, 171, 39}  
\definecolor{color8}{RGB}{161, 16, 53}  
\definecolor{color9}{RGB}{189, 205, 0} 
\definecolor{color10}{RGB}{97, 33, 88}   
\definecolor{color11}{RGB}{255, 237, 0} 
\definecolor{color12}{RGB}{122, 111, 172} 

\definecolor{lightgray}{RGB}{230,230,230}
\definecolor{gray}{RGB}{180,180,180}
\definecolor{darkgray}{RGB}{70,70,70}

\definecolor{rwthLightBlue}{RGB}{142, 186, 229}
\definecolor{rwthLightGreen}{RGB}{184, 214, 152}

\begin{document}

\makeatletter
\def\ps@IEEEtitlepagestyle{%
  \def\@oddfoot{\mycopyrightnotice}%
  \def\@evenfoot{}%
}
\def\mycopyrightnotice{%
    \begin{minipage}{\textwidth}
\centering \scriptsize
This work has been submitted to the IEEE for possible publication. Copyright may be transferred without notice, after which this version may no longer be accessible.\hfill% <--- Change here
    \end{minipage}
    \gdef\mycopyrightnotice{}% just in case
}
\makeatother

%%%%%%%%%%%%%%%%%%%%%%%%%%%%%%%%%%%%%%%%%%%%%%%%%%%%%%
\title{
Ufil: A Unified Framework for Infrastructure-based Localization
\thanks{
$^{1}$ The authors are with the Chair of Embedded Software, RWTH Aachen
University, {\texttt{\{lastname\}@embedded.rwth-aachen.de}}

$^{2}$ B. Alrifaee is with the Department of Aerospace Engineering, University of the Bundeswehr Munich, Germany, {\texttt{bassam.alrifaee@unibw.de}}

$^{\dagger}$ Equal contribution

We acknowledge the financial support for this project by the Collaborative Research Center / Transregio 339 of the German Research Foundation (DFG).
}
}

\author{
Simon Schäfer$^{1}$\,\orcidlink{0000-0002-6482-2383}, Lucas Hegerath$^{1,\dagger}$\,\orcidlink{0000-0003-2926-2664}, Marius Molz$^{1,\dagger}$\,\orcidlink{0009-0005-8406-8727},~\IEEEmembership{Graduate~Student~Members,~IEEE}, \\Massimo Marcon$^{1}$\,\orcidlink{0009-0009-0876-6573}, and Bassam Alrifaee$^{2}$\,\orcidlink{0000-0002-5982-021X},~\IEEEmembership{Senior Member,~IEEE}
}
%%%%%%%%%%%%%%%%%%%%%%%%%%%%%%%%%%%%%%%%%%%%%%%%%%%%%%

\maketitle

\begin{abstract}
Infrastructure-based localization enhances road safety and traffic management by providing state estimates of road users. Development is hindered by fragmented, application-specific stacks that tightly couple perception, tracking, and middleware. We introduce Ufil, a Unified Framework for Infrastructure-Based Localization with a standardized object model and reusable multi-object tracking components. Ufil offers interfaces and reference implementations for prediction, detection, association, state update, and track management, allowing researchers to improve components without reimplementing the pipeline. Ufil is open-source C++/ROS~2 software with documentation and executable examples. We demonstrate Ufil by integrating three heterogeneous data sources into a single localization pipeline combining
\begin{enumerate*}
    \item vehicle onboard units broadcasting ETSI ITS-G5 Cooperative Awareness Messages,
    \item a lidar-based roadside sensor node, and
    \item an in-road sensitive surface layer.
\end{enumerate*}
The pipeline runs unchanged in the CARLA simulator and a small-scale CAV testbed, demonstrating Ufil’s scale-independent execution model. In a three-lane highway scenario with \num{423} and \num{355} vehicles in simulation and testbed, respectively, the fused system achieves lane-level lateral accuracy with mean lateral position RMSEs of \qty{0.31}{\meter} in CARLA and \qty{0.29}{\meter} in the CPM Lab, and mean absolute orientation errors around \qty{2.2}{\degree}. Median end-to-end latencies from sensing to fused output remain below \qty{100}{\milli\second} across all modalities in both environments.
\end{abstract}

%%%%%%%%%%%%%%%%%%%%%%%%%%%%%%%%%%%%%%%%%%%%%%%%%%%%%%

%%%%%%%%%%%%%%%%%%%%%%%%%%%%%%%%%%%%%%%%%%%%%%%%%%%%%%
\begin{IEEEkeywords}
Infrastructure-based localization, Object Detection and Tracking, Traffic Monitoring, Simulation, Small-scale Testbed
\end{IEEEkeywords}
%%%%%%%%%%%%%%%%%%%%%%%%%%%%%%%%%%%%%%%%%%%%%%%%%%%%%%

%%%%%%%%%%%%%%%%%%%%%%%%%%%%%%%%%%%%%%%%%%%%%%%%%%%%%%
\section*{Open Material}
\label{sect:openmaterial}
%%%%%%%%%%%%%%%%%%%%%%%%%%%%%%%%%%%%%%%%%%%%%%%%%%%%%%

\begin{flushleft}
\begin{tabular}{@{}L{1cm} L{7.2cm}}
\textbf{Code} & \href{https://github.com/bassamlab/Ufil}{github.com/bassamlab/Ufil} \\
\textbf{Docs} & \href{https://github.com/bassamlab/Ufil-docs}{github.com/bassamlab/Ufil-docs} \\
\end{tabular}
\end{flushleft}

%%%%%%%%%%%%%%%%%%%%%%%%%%%%%%%%%%%%%%%%%%%%%%%%%%%%%%
\section{Introduction}
\label{sect:intro}
%%%%%%%%%%%%%%%%%%%%%%%%%%%%%%%%%%%%%%%%%%%%%%%%%%%%%%

\subsection{Motivation}
\label{sect:intro_motivation}

Vehicle-centric perception has been shown to be insufficient for achieving reliable, large-scale coordination in complex traffic environments. Limitations in sensing range, occlusions, and incomplete situational awareness restrict the ability of individual vehicles to form a consistent representation of the surrounding traffic scene. Infrastructure-based localization of road users refers to the process of using sensors embedded in roads and the surrounding infrastructure to estimate the states of all road users within a fixed region of interest. This approach promises to enhance safety by sharing state estimations with \acp{CAV}, as well as reduce traffic congestion by providing data for more efficient trajectory planning. However, developing such systems in real-world scenarios often involves long development cycles, high costs, and regulatory hurdles related to safety and privacy. To address these challenges, developers typically create simulations of proposed systems before deployment, ensuring satisfactory results are achieved prior to real-world implementation.

Several frameworks exist for simulating and developing these systems. Autoware is an open-source software stack designed for self-driving vehicles that supports various functionalities but lacks a dedicated focus on infrastructure-centric applications \cite{kato2018autoware}. It has been utilized in test vehicles such as Edgar \cite{karle2024edgarautonomousdrivingresearch}. The MATLAB Automated Driving Toolbox also serves as a comprehensive environment for designing automated driving algorithms; however, it primarily emphasizes vehicle-centric models \cite{mathworks2025automated}. While it is possible to develop an infrastructure-based perception system within these frameworks, they predominantly cater to automated driving contexts that require defining an ego vehicle around which the coordinate system centers, an inconvenient requirement for fixed perspectives typical of infrastructure-based systems.

Moreover, while simulations can provide rough estimates of final products, transitioning from simulation to real-world application introduces additional challenges that cannot be fully replicated in virtual environments; this phenomenon is commonly referred to as the sim-to-real gap. In recent years, there has been growing interest in small-scale testbeds that consist of scaled-down versions of real-world traffic settings, including physical vehicles and road networks, alongside distributed computation in networked environments \cite{mokhtarian2024survey}. By integrating physical components with networks, such testbeds offer scenarios that closely resemble real-world deployments while effectively reducing the sim-to-real gap \cite{schaefer2024from}.

Recent work has surveyed simulators for automated driving; it found that CARLA is the only open-source simulator that provides all necessary components for simulating an infrastructure-based localization system \cite{yueyuan2024choose}. CARLA allows simulation of traffic dynamics along with static/dynamic infrastructure and sensor data across common sensor technologies. Recent extensions include CARLOS, which enables large-scale testing with enhanced \ac{ROS 2} integration \cite{geller2024carlos}.

Various datasets relevant to infrastructure localization exist, but do not provide general frameworks or tooling support. Examples include the \emph{A9-Dataset}, featuring manually labeled cameras and lidar data for machine learning \cite{cress2022a9}, the \emph{LUMPIE} Dataset, which also contains manually labeled cameras and lidar data \cite{busch2022lumpi}, and \emph{UrbanIng-V2X}, which follows similar labeling efforts \cite{urbaningv2x2025}.

Other notable frameworks with open code include \emph{CoCa3D}, a camera-only framework \cite{yu2023coca}. Additionally, \emph{DAIR-V2X/OpenDAIRV2X} \cite{haibao2022dairv2x} focuses on Vehicle-to-Infrastructure communication while providing datasets aimed at machine learning models. The \emph{INFRA-3DRC} dataset presents a sensor setup with radar and camera alongside the dataset and provides tracking code for its specific scenario \cite{agrawal2024semi}. In \emph{CoopScenes}, the authors fuse one sensor node with vehicle data, providing modality-specific tracking code written in Python, which cannot be easily generalized \cite{vosshans2025coop}. \emph{FlexSense} presents sensing system details along with published datasets but lacks operational code. Some works, like \cite{polley202525dobjectdetectionintelligent}, publish functioning machine learning models without providing comprehensive frameworks for them. More complete works, such as \emph{SetThmSLAM} \cite{li2023set}, provide a full software suite for robustly localizing autonomous vehicles using infrastructure sensors but are limited by their implementation details. \emph{SetThmSLAM} is written in the proprietary MATLAB system, without the possibility of abstraction for other use cases. Lastly, \emph{OpenCOOD} officially listed 12 supported projects but stopped being maintained in 2023; it was designed as an Open COOperative Detection framework for autonomous driving. The main focus of \emph{OpenCOOD} is the training of machine learning algorithms and in combination with \emph{OpenCDA} mainly targeting offline cooperative perception. More up-to-date examples that still utilize \emph{OpenCOOD} include \emph{HEAL} (HEterogeneous ALliance) \cite{lu2024an}, which supports lidar and camera integration as well as the previously mentioned \emph{UrbanIng-V2X} dataset.

Despite advancements made within these frameworks, three significant gaps remain: 
\begin{enumerate*}
    \item there is currently no framework explicitly targeting infrastructure-based localization that provides all necessary components while simultaneously leveraging opportunities presented by the static nature of the problem;
    \item limited research exists focused on utilizing small-scale testbeds as intermediate stages for validating such systems before transitioning them into real-world environments; finally, 
    \item there is no existing framework designed for agnostic execution allowing scale-independent testing of localization pipelines against simulations, small-scale testbeds, and full-scale deployments.
\end{enumerate*}

\subsection{Contribution}
\label{sect:intro_contribution}

To address these gaps, this paper proposes a \emph{\ac{Ufil}}, an open-source framework that provides essential building blocks for developing and validating infrastructure-based perception systems. \ac{Ufil} aims to be data-source-agnostic and scenario-independent.
Concretely, we contribute:
\begin{itemize}
    \item a unified, history-aware object model and modular tracking pipeline tailored to infrastructure-based localization,
    \item reference implementations and interfaces for prediction, detection, association, update, and management, enabling plug-and-play comparison of algorithms, and
    \item a scale-independent execution model demonstrated on a single highway perception pipeline that runs unchanged in CARLA and in a small-scale \ac{CAV} testbed.
\end{itemize}
We present this paper in connection with the open-source code base, documentation, and application examples.

\subsection{Outline}
\label{sect:intro_outline}

The structure of this paper is organized as follows: In \Cref{sect:framework}, we present the \ac{Ufil} framework and explain how its design choices address the gaps identified above. \Cref{sect:example} then instantiates the framework in a concrete infrastructure-based localization system for a three-lane highway, which serves as a case study. \Cref{sect:results} details our evaluation methodology and reports tracking accuracy and latency for this system in CARLA and in the \ac{CPM Lab} testbed. Finally, \Cref{sect:conclusion} concludes our findings and discusses potential directions for future work.

%%%%%%%%%%%%%%%%%%%%%%%%%%%%%%%%%%%%%%%%%%%%%%%%%%%%%%
\section{Notation}
\label{sect:notation}
%%%%%%%%%%%%%%%%%%%%%%%%%%%%%%%%%%%%%%%%%%%%%%%%%%%%%% 

In this paper, we adopt the following notation conventions: 

\begin{itemize}
    \item Lowercase bold letters, such as $\pmb{a}$, represent vectors, while uppercase bold letters, such as $\pmb{A}$, denote matrices. Calligraphic letters, like $\mathcal{A}$, denote sets.
    \item The superscript $\left(.\right)^T$ indicates the transpose of a vector or matrix, and $\left(.\right)^{-1}$ represents the inverse of a matrix. An estimate of the true value $\left(.\right)$ is denoted by $\hat{\left(.\right)}$.
    \item The subscript $\left(.\right)_k$ specifies the value at time step $k$, and $\left(.\right)_{j,k}$ refers to the $j$-th value at time step $k$. Additionally, $\left(.\right)^i$ designates the $i$-th component of a selected set, where $i \in \{a, b, c\}$.
\end{itemize}

%%%%%%%%%%%%%%%%%%%%%%%%%%%%%%%%%%%%%%%%%%%%%%%%%%%%%%
\section{Framework}
\label{sect:framework}
%%%%%%%%%%%%%%%%%%%%%%%%%%%%%%%%%%%%%%%%%%%%%%%%%%%%%%

This section introduces \emph{\ac{Ufil}}, which forms the core of the work presented in the remainder of the paper. 
We first summarize the overall concept, including the object model and design goals (\Cref{sect:framework_concept}). 
We then detail each step of the multi-object tracking cycle, showing how prediction, detection, association, update, and management are exposed as modular, exchangeable components (\Cref{sect:framework_prediction}–\ref{sect:framework_management}). 
We conclude with a simple simulation tool for exploring design choices (\Cref{sect:framework_simple_simulation}), which complements the system-level evaluation in \Cref{sect:example,sect:results}.

\subsection{Concept}
\label{sect:framework_concept}

\ac{Ufil} is a collection of reusable software components that offer standardized definitions, algorithms, and message formats for infrastructure-based localization and tracking applications. 
Its design is guided by three principles derived from \Cref{sect:intro}: (i) separate generic tracking logic from application-specific processing, (ii) make all major algorithmic choices pluggable, and (iii) support the same pipelines across simulation, small-scale testbeds, and real-world deployments.

At the core of \ac{Ufil} is a C++ header-only library that provides the unified object model and middleware-independent algorithms, including motion models, association metrics, and state update routines. 
The surrounding \ac{ROS 2} packages integrate this core library into a distributed computation architecture, handle configuration and parameter management, and provide tooling for logging, visualization, and evaluation. 
While the core library can be used without \ac{ROS 2}, the integration layer implements most of the functionality required for real deployments and large-scale experiments.

\begin{figure}[t]
    \centering
    \includegraphics{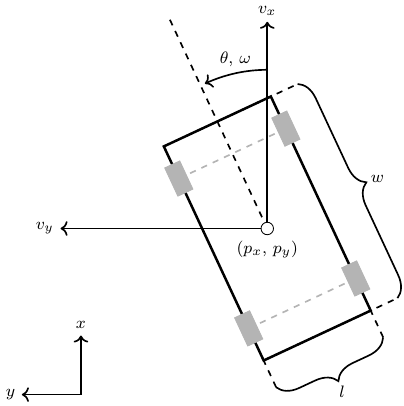}
    \caption{Object definition in a bird’s eye view.}
    \label{fig:object_definition}
\end{figure}

\ac{Ufil} represents all road users with a unified object model, illustrated in \Cref{fig:object_definition}. 

Each object $\object_{i,k}$ is defined as
\begin{equation}
    \object_{i,k} = \{\estimatedstatevector, \statecovariancematrix, \estimateddimensionvector, \dimensionvariancematrix, \existenceprobability, \classificationvector \},
    \label{eq:object}
\end{equation}
where $\estimatedstatevector \in \mathbb{R}^6$ is the estimated state, $\statecovariancematrix \in \mathbb{R}^{6 \times 6}$ the state covariance, $\estimateddimensionvector \in \mathbb{R}^3$ and $\dimensionvariancematrix \in \mathbb{R}^{3 \times 3}$ the dimensions and their covariance, $\existenceprobability \in [0,1]$ the existence probability, and $\classificationvector \in [0,1]^7$ the class probability vector with
\begin{equation}
    \sum_{j=1}^{7} \classificationvector^{(j)} = 1.
\end{equation}
The index $i$ denotes the $i$th element of the current object list $\currentobjectlist$, and $k$ the time step. By default, we use seven classes
\begin{equation}
\{\text{car}, \text{truck}, \text{motorcycle}, \text{bicycle}, \text{pedestrian}, \text{stationary}, \text{other}\},
\end{equation}
in which the last class represents uncertainty and allows for the communication of the absence of a classification result.

The state vector encapsulates position and motion data as $\statevector = \begin{bmatrix}\positionvector & \velocityvector & \yaw & \yawrate\end{bmatrix}^T,$
with $\positionvector = \begin{bmatrix}\positionx & \positiony\end{bmatrix}^T \in \mathbb{R}^2$, $\velocityvector = \begin{bmatrix}\velocityx & \velocityy\end{bmatrix}^T \in \mathbb{R}^2$,
yaw $\yaw \in [0,360)$, and yaw rate $\yawrate \in \mathbb{R}$.
The dimension vector represents the length, width, and hight of the object enclosing bounding box as $\dimensionvector = \begin{bmatrix}\length & \width & \height\end{bmatrix}^T \in \mathbb{R}^3$

For each discrete time step $k$, the tracker stores the full object list $\currentobjectlist = \{\object_1, \dots, \object_N \}$, and each object retains a time-stamped sequence of its history. 
This history supports retrospective association and moving-horizon estimation. 
The maximum history length is controlled by pruning policies in the management step (\Cref{sect:framework_management}).

To support reduced state representations, \ac{Ufil} requires at least position, length, and width. 
If a tracking algorithm uses only a subset of the state space, the unused components are indicated to downstream components. In the \texttt{ufil\_ros} package, they are marked in the serialized covariance matrix by inserting $-1$ on the corresponding diagonal entries. Library-level operations never use these entries directly; instead, selection matrices and masks ensure that only modeled components are processed. This convention allows downstream components to work on partial states without changing interfaces while avoiding misuse of non-physical covariance values.

\ac{Ufil} follows the standard multi-object tracking cycle
Prediction $\rightarrow$ Detection $\rightarrow$ Association $\rightarrow$ Update $\rightarrow$ Management, 
and provides interchangeable algorithms for each step. 
All steps share a common interface and operate on the unified object model, which simplifies algorithm exchange and comparison and underpins the portability demonstrated later in \Cref{sect:example,sect:results}.

\subsection{Prediction}
\label{sect:framework_prediction}

\ac{Ufil} applications implement discrete-time multi-object tracking. 
For each time step $k$, the prediction step computes the most probable state of each object at time step $k$ given its state at time step $k-1$ and a motion model.

Prediction is structured around the four main object components: kinematic state $\estimatedstatevector$, dimensions $\estimateddimensionvector$, classification $\classificationvector$, and existence probability $\existenceprobability$. 
While motion models for the prediction of the kinematic state are well established, there is no de facto standard for the prediction of dimensions, classification, or existence, so \ac{Ufil} provides simple defaults which keep these values constant and interfaces for custom predictors.
Out of the box, \ac{Ufil} supports six motion models: No-Prediction, Random Walk, Constant Velocity, Extended Constant Velocity, Constant Acceleration, and Extended Constant Acceleration, all based on \cite{schubert2008comparison}. 
These can be implemented as linear models, but \ac{Ufil} also supports non-linear transitions via a generic interface.

For linear models, prediction follows a Kalman-style update:
\begin{equation}
    \estimatedstatevector_{k}^{-} = \statetransitionmatrix_{k}\,\estimatedstatevector_{k-1} + \stateinputmatrix_{k}\,\stateinput_{k} + \omega_{k},
\end{equation}
where $\statetransitionmatrix_{k}$ is defined by the motion model, and $\stateinputmatrix_{k}, \stateinput_{k}$ encode known inputs, typically omitted in infrastructure-based tracking.

With $\omega_{k} \sim \mathcal{N}(0,\processnoisematrix_{k})$, the covariance prediction is
\begin{equation}
 {{\statecovariancematrix}_{,k}}^{-} = \statetransitionmatrix_{k}\,{\statecovariancematrix}_{,k-1}\,\statetransitionmatrix_{k}^{T} + \processnoisematrix_{k},
\end{equation}
where $\processnoisematrix_{k}$ is modeled as integrated white noise based on \cite{barshalom2002estimation}.

\subsection{Detection}
\label{sect:framework_detection}

\ac{Ufil} operates on object-level data, but many sensors provide raw data (e.g., point clouds or images). 
A detection step is therefore required to convert raw measurements into object hypotheses.

Unlike architectures that strictly separate detection and tracking, \ac{Ufil} integrates detection into the tracking pipeline. 
Detectors can use predicted objects from the previous time step as priors, for example, to select regions of interest, and can pre-populate parts of the association matrix $\currentassociationmatrix$.

\ac{Ufil} ships with two implemented detectors. 
The first is a pass-through detector that forwards inputs already at the object level. 
The second is a lidar-based detector presented later in \Cref{sect:example}. 
For any custom detector, the framework provides an open interface.

\subsection{Association}
\label{sect:framework_association}

Association finds correspondences between detections at time step $k$ and predicted objects. 
\ac{Ufil} formulates this as an assignment problem with cost matrix $\currentcostmatrix$, where each entry encodes the cost of pairing a detection with an object.

\ac{Ufil} provides four cost measures:
\begin{enumerate*}
    \item Euclidean distance,
    \item Mahalanobis distance,
    \item closed-form Wasserstein distance, and
    \item bounding box Intersection over Union (IoU).
\end{enumerate*}
User-defined measures can be added via a generic interface.

The cost matrix is passed to an assignment solver, which produces an association matrix $\currentassociationmatrix$. 
\ac{Ufil} implements:
\begin{enumerate*}
    \item the Linear Assignment Problem Jonker--Volgenant algorithm (LAPJV),
    \item a simplex-based linear programming solver,
    \item a greedy associator, and
    \item a Hungarian associator.
\end{enumerate*}
Optionally, gating discards assignments above a configurable cost threshold $\costmatrixthreshold$; for Mahalanobis distances, $\costmatrixthreshold$ can be chosen as a $\chi^2$-based threshold, while for other metrics it is specified in metric units. 
The resulting Boolean association matrix $\currentassociationmatrix$ is used to update objects and identify unassigned detections.

\begin{figure}[t]
    \centering
    \includegraphics{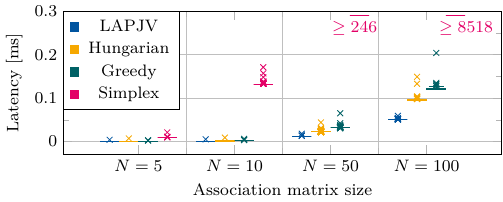}
    \caption{Runtimes of the four solvers implemented in \ac{Ufil}. Each solver solved the linear assignment problem 100 times for a random $N \times N$ matrix.}
    \label{fig:association_evaluation} 
\end{figure}

A key advantage of integrating multiple solvers into a single framework with a unified interface is that they can be easily compared under identical conditions. 
For a fixed cost matrix and gating setup, the Hungarian, LAPJV, and simplex-based solvers compute the same optimal one-to-one assignment (up to numerical tolerances), while the greedy associator is faster, but not guaranteed to be globally optimal. 
We present a runtime benchmark in \Cref{fig:association_evaluation}, where each solver solves a linear assignment problem \num{100} times for growing matrix sizes $N$. 
The results show that the Hungarian and LAPJV algorithms yield optimal assignments with consistently low runtimes, whereas the simplex-based solver exhibits significantly higher runtimes than the other methods.

\subsection{Update}
\label{sect:framework_update}

The update step incorporates the associated detections into the object list and refines the estimates. 
Analogous to prediction, \ac{Ufil} conceptually distinguishes updates for kinematic state, dimensions, classification, and existence probability.

The kinematic state is updated using linear or nonlinear Kalman-filter-based updaters. 
For an associated prediction–measurement pair, the innovation and its covariance are
\begin{align}
\innovationvector_{k} &= \measurementvector_{k} - \measurementmatrix_{k}\,\estimatedstatevector^{-}_{k}, \\
\innovationmatrix_{k} &= \measurementmatrix_{k}\,{{\statecovariancematrix}_{,k}}^{-}\,\measurementmatrix_{k}^{T} + \measurementnoisematrix_{k}.
\end{align}
Using the Joseph form \cite{bucy1987filtering} for numerical stability, \ac{Ufil} updates state and covariance:
\begin{align}
\pmb{K}_{k} &= {{\statecovariancematrix}_{,k}}^{-}\,\measurementmatrix_{k}^{T}\,\innovationmatrix_{k}^{-1},\\
\estimatedstatevector_{k} &= \estimatedstatevector_{k}^{-} + \pmb{K}_{k}\,\innovationvector_{k},\\
{{\statecovariancematrix}_{,k}} &= (I - \pmb{K}_{k}\,\measurementmatrix_{k})\,{{\statecovariancematrix}_{,k}}^{-}\,(I - \pmb{K}_{k}\,\measurementmatrix_{k})^{T} + \pmb{K}_{k}\,\measurementnoisematrix_{k}\,\pmb{K}_{k}^{T}.
\end{align}
The measurement matrix $\measurementmatrix_{k}$ is built per measurement to select observed state components.

\ac{Ufil} provides the following out-of-the-box updaters and interfaces for dimension, classification, and existence:
\begin{enumerate*}
    \item a grid-map estimator based on \cite{aeberhard2017object} using occupancy to estimate the dimension,
    \item a dimension-based classifier presented in \cite{schaefer2025lidar} using road-user statistics, and
    \item a heuristic existence update from \cite{schaefer2025lidar} and a Bayesian estimator from \cite{aeberhard2017object}.
\end{enumerate*}
All updaters provide an open interface and can be replaced by user-specific algorithms. Detailed parameter choices, priors for the dimension-based classifier, and existence-model settings are documented alongside the \ac{Ufil} examples.

\subsection{Management}
\label{sect:framework_management}

The management step creates new objects, deletes obsolete ones, and prunes histories to keep the object history consistent and reduce computational load.

\ac{Ufil} provides two deletion strategies:
\begin{enumerate*}
    \item \emph{time-based:} remove objects without associated detections for longer than a threshold and
    \item \emph{existence-based:} remove objects with existence probability below a threshold.
\end{enumerate*}

New objects are initialized from unassigned detections using a user-provided function that sets initial state and covariance, as well as dimension, classification, and initial existence probability. \ac{Ufil} also supports tentative tracks and $M/N$ confirmation logic to avoid promoting short-lived false positives to confirmed tracks.

History pruning is available in three modes: no pruning, time-based pruning, and pruning by number of entries, the latter being useful for moving-horizon estimators. 
This allows trading off temporal context against memory and runtime.

\subsection{Simple Simulation}
\label{sect:framework_simple_simulation}

Selecting a suitable combination of motion models, association metrics, solvers, and management strategies is non-trivial, as interactions between components and runtime effects are often hard to predict. To support such design decisions, \ac{Ufil} provides a simple scenario simulator in the \texttt{ufil\_examples} package that focuses on the multi-object tracking system and explicitly excludes the detection step by providing object-level detections. Users can specify short, controlled scenarios via configuration files and benchmark different tracking configurations against each other.

As an illustrative case, we consider two pedestrians following trajectories that intersect at a crossing point, a classic situation in which trackers tend to swap identities. Around the intersection, the positions become very similar, and association is particularly challenging. We run a fixed prediction–update–management chain and only vary the association stage by combining different solvers with Euclidean and Wasserstein distance hypotheses. In this scenario, the choice of solver (Hungarian, Greedy) does not affect accuracy, as they compute the same optimal assignment for a given cost matrix. In contrast, the hypothesis function has a strong effect: Euclidean distance fails to maintain identities near the intersection, while Wasserstein distances exploit additional state information to resolve the ambiguity. While the greedy solver provides a fast, though not always optimal, baseline it does perform well in such simple scenarios. This example highlights how the simulator can be used to explore algorithmic trade-offs before deployment and complements the system-level experiments in \Cref{sect:example} and \Cref{sect:results}.

\begin{figure}[t]
    \centering
    \subfloat[Results with euclidean hypothesizer.\label{fig:pedestrian_simulation_results_euclidean}]{%
       \includegraphics{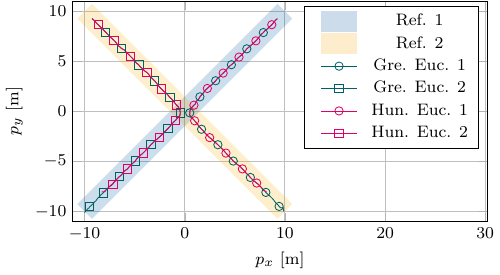}
    }
    \\
    \subfloat[Results with wasserstein hypothesizer.\label{fig:pedestrian_simulation_results_wasserstein}]{%
        \includegraphics{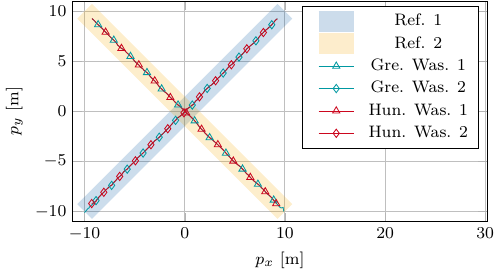}
    }
    \caption{Results of the simple simulation crossing example for different hypothesizer  and association solvers.}
    \label{fig:pedestrian_simulation_results}
\end{figure}

%%%%%%%%%%%%%%%%%%%%%%%%%%%%%%%%%%%%%%%%%%%%%%%%%%%%%% 
\section{Example}
\label{sect:example}
%%%%%%%%%%%%%%%%%%%%%%%%%%%%%%%%%%%%%%%%%%%%%%%%%%%%%% 

The previous section introduced \ac{Ufil}’s building blocks and interfaces. 
We now show how these blocks can be instantiated into a concrete infrastructure-based localization system, and we use this system in \Cref{sect:results} for experimental evaluation.
The example system is not intended as a new tracking algorithm; it serves as a case study demonstrating how \ac{Ufil} supports heterogeneous sensors and execution across different environments.
Additional modalities (e.g., radar, cameras, other \ac{V2I} messages) can be integrated via detectors that produce \ac{Ufil} objects.

We begin with the overall concept of the system in \Cref{sect:example_concept}. 
We then discuss its integration in a simulator in \Cref{sect:example_high_fidelity_simulation} and its deployment in a small-scale testbed in \Cref{sect:example_testbed}. 

\subsection{Concept}
\label{sect:example_concept}

The proposed infrastructure-based perception system collects data from three complementary sources. 
The goal is to cover a representative set of modalities routinely discussed in the literature while showcasing the versatility of the framework:
\begin{itemize}
    \item \textbf{\ac{V2I} communication:} \ac{CAM} as defined in ETSI EN\,302\,637-2, a C-V2X standard already used on German roads \cite{kueppers2024v2aix}. 
    \ac{Ufil} fully integrates this standard, including parsing \ac{CAM} messages, converting them into the unified object model, and generating standard-compliant messages with correct timing.
    \item \textbf{Roadside lidar:} a lidar-based \ac{OSN} performing detection and tracking, implemented according to \cite{schaefer2025lidar}.
    \item \textbf{In-road sensing:} a \ac{SSL} system as investigated in \cite{schaefer2023investigating}, providing object-level estimates including axle and wheel information.
\end{itemize}

Each data source delivers an object list $\objectlist$ to an \ac{RSU} running a centralized fusion pipeline composed of \ac{Ufil} components. 
The \ac{RSU} fuses all lists and outputs a single global object list for the observed road segment. 
The \ac{RSU} fuses all lists and outputs a single global object list for the observed road segment. 

The specific measurement models, calibration and synchronization procedures, and modality-specific parameters are described in the \ac{Ufil} documentation and example repositories, allowing this paper to focus on the framework rather than on sensor specifics.

For testing such pipelines, \ac{Ufil} provides two testing approaches out of the box: support for the high-fidelity simulator CARLA \cite{dosovitskiy2017carla} and for the small-scale cyber-physical testbed \ac{CPM Lab} \cite{kloock2021cyberphysical}. 
Additionally, the lidar-based tracker in \cite{schaefer2025lidar} already used an early version of \ac{Ufil} in a real-world deployment, providing first evidence that \ac{Ufil}’s abstractions carry over to real infrastructure. 
A full replication of the three data sources presented above in the real world is currently limited by the lack of a full-scale in-road sensing system; the software stack itself is ready for such an extension.

\subsection{High Fidelity Simulation}
\label{sect:example_high_fidelity_simulation}

CARLA and \ac{Ufil} both provide \ac{ROS 2} interfaces, enabling an easy connection. 
An adapter node subscribes to CARLA topics (e.g., ground-truth vehicle states) and publishes \ac{Ufil} object lists. 
These ground-truth lists are the input for three downstream applications. First, they drive the \ac{SSL} simulation, implemented as a co-simulation. Second, they provide data for the generation of the \ac{V2I} messages, and lastly, they serve as ground truth for evaluation. 
The lidar point clouds come directly from CARLA’s sensor models and are processed by the lidar-based \ac{OSN} module without further adaptation.

Scenarios in CARLA are configured via configuration files that specify maps, sensor poses, and traffic spawn/deletion points. 
\ac{Ufil} ships with ready-made configurations for one intersection in \texttt{Town10} and a highway segment in \texttt{Town05}, the latter matching the scenario used later in \Cref{sect:results}.

\subsection{Small-scale testbed}
\label{sect:example_testbed}

The \ac{CPM Lab} \cite{kloock2021cyberphysical} is a small-scale \ac{CAV} testbed featuring physical vehicles, static infrastructure, and a distributed computation approach. 
We retrofit it with a \emph{Sensing Tex} pressure-sensitive mat representing the in-road sensing system, a roadside \emph{Intel RealSense L515} lidar sensor representing the lidar, and \ac{V2I} capabilities for the existing small-scale vehicles called \emph{\textmu Cars} \cite{scheffe2020networked}.

An adapter node converts the lab’s internal vehicle state messages into \ac{Ufil} object lists, analogous to the CARLA adapter. 
Lidar and \ac{SSL} sensors are connected via \ac{ROS 2} driver nodes that publish \ac{ROS 2} standard messages, which are directly consumed by \ac{Ufil} components. 
Because each sensor runs on its own compute unit and communicates via the testbed network, realistic effects such as communication latency and clock offsets are represented faithfully and propagate through the same pipeline used in CARLA.

All configurations, launch files, and drivers are provided in the \texttt{ufil\_examples} package and documented in the \ac{Ufil} documentation.

%%%%%%%%%%%%%%%%%%%%%%%%%%%%%%%%%%%%%%%%%%%%%%%%%%%%%% 
\section{Experimental results and discussion}
\label{sect:results}
%%%%%%%%%%%%%%%%%%%%%%%%%%%%%%%%%%%%%%%%%%%%%%%%%%%%%% 

This evaluation aims to provide performance insights into the presented system from \Cref{sect:example} and evaluate the \ac{Ufil}-based perception pipeline in both CARLA and the \ac{CPM Lab}. 
The goal is to assess whether a single framework-enabled application can achieve comparable accuracy and latency even when tested in two separate environments. Rather than optimizing a particular tracking algorithm, we focus on demonstrating \ac{Ufil}’s scale-independent execution.

For offline evaluation, we record all object lists as \ac{ROS 2} bag files and process them with \ac{Ufil}'s evaluation scripts as described below.

\subsection{Setup}
\label{sect:results_setup}

The evaluation employs the highway perception system from \Cref{sect:example} as a case study for \ac{Ufil}’s scale-independent execution. The scenario is a three-lane highway segment with an \ac{SSL} covering the two right-most lanes near the upstream part of the segment and a roadside lidar \ac{OSN} installed downstream, facing the \ac{SSL} region; their fields of view partially overlap. Vehicles are equipped with \ac{V2I} functionality and broadcast \ac{CAM} messages. The same configuration of motion models, association settings, and management policies is used in all experiments.

The scenario is instantiated in CARLA and in the CPM Lab. Each experiment runs for approximately \qty{15}{\minute} and contains only cars as moving objects. Over one run, CARLA comprises \num{423} unique vehicles and the CPM Lab \num{355} vehicles, with typically five to six cars present in the field of view at any given time. CAM messages follow the ETSI update rules with an effective rate between \qtyrange{1}{10}{\hertz} depending on driving state, while lidar and \ac{SSL} operate at \qty{10}{\hertz}. The centralized fusion node on the \ac{RSU} runs as a time-triggered system at \qty{50}{\hertz}.

For offline evaluation, the ground-truth and fused object lists are recorded as \ac{ROS 2} bag files and converted into HDF5 files containing all object data separated by IDs. \ac{Ufil}’s evaluation scripts then align reference and estimated trajectories in the local, georeferenced Cartesian map frame used by all components. For each experiment, candidate pairs of reference and estimated tracks are formed and, for each pair, the position \ac{RMSE} over the joint lifetime is computed. Pairs with \ac{RMSE} above \qty{3}{\meter} are discarded as mismatches; the remaining pairs are selected greedily by increasing \ac{RMSE}, and unmatched estimates are treated as ghost tracks. The resulting matched tracks are used to compute longitudinal, lateral, and orientation errors over time.

\begin{figure}[t]
    \centering
    \includegraphics{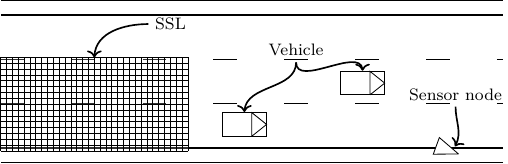}
    \caption{Testing scenario: a three-lane highway segment with \ac{SSL}, lidar \ac{OSN}, and \ac{V2I}-equipped vehicles.}
    \label{fig:testing_scenario}
\end{figure}

\subsection{Track Accuracy}
\label{sect:results_track_accuracy}

We run the \ac{Ufil}-based perception pipeline on the CARLA and \ac{CPM Lab} recordings and process the resulting bag files with the evaluation scripts described above. For each matched trajectory pair, we compute errors in longitudinal position $\positionx$, lateral position $\positiony$, and orientation $\yaw$, and then aggregate across all tracks in the scenario.

In both environments, the fused system achieves lane-level lateral accuracy across the area of interest. Errors are smallest in the region where \ac{SSL}, lidar, and \ac{V2I} overlap, due to redundant observations, and increase slightly outside the \ac{SSL} coverage where lidar and \ac{CAM} dominate. For the CARLA scenario, the mean longitudinal and lateral position \acp{RMSE} are \qty{0.11}{\meter} and \qty{0.31}{\meter}, respectively. For the CPM Lab, they are \qty{0.29}{\meter} in both directions. Given a typical lane width of approximately \qty{3.5}{\meter}, the lateral errors are well within lane-level accuracy. The mean absolute orientation error is \qty{2.12}{\degree} in CARLA and \qty{2.36}{\degree} in the CPM Lab. 

\begin{figure}[t]
    \centering
    \subfloat[Tracking error in simulation.\label{fig:tracking_accruacy_simulation}]{%
       \includegraphics{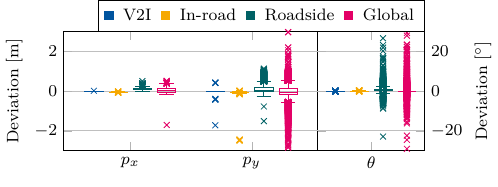}
    }
    \\
    \subfloat[Tracking error in small-scale testbed.\label{fig:tracking_accruacy_small_scale}]{%
        \includegraphics{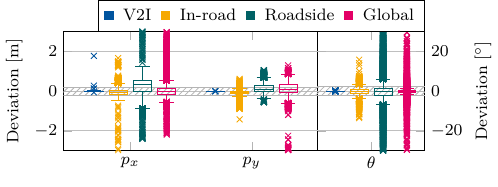}
    }
  \caption{Summary of state accuracy. For each domain, we plot the error for positions $\positionx$, $\positiony$ (left) and orientation $\yaw$ (right). For the small-scale testbed we indicate the accuracy of the reference system in gray.}
  \label{fig:summary_state_accuracy} 
\end{figure}

The CPM Lab uses a reference system rather than a ground-truth system; thus, it provides a basis for computing errors relative to its mean position accuracy of approximately \qty{19.89}{\centi\meter} and a mean absolute orientation error of \qty{2.25}{\degree} \cite{kloock2020vision}. Consequently, any computed errors that fall below these reference values may not be meaningful, as they could be indicative of inaccuracies within the reference system itself rather than true performance metrics. We indicated this reference accuracy in the \Cref{fig:tracking_accruacy_small_scale}.

Qualitatively, the error distributions exhibit similar trends in both environments: longitudinal errors remain small over the full segment, while lateral errors increase slightly in regions with oblique sensor views or reduced overlap between modalities. This indicates that the small-scale testbed approximates full-scale behavior well and that \ac{Ufil}’s abstractions do not introduce environment-specific degradation.

\subsection{End-to-End Latency}
\label{sect:results_latency}

To assess timing behavior, \ac{Ufil} propagates sensor timestamps through the pipeline. Each fused object list emitted by the \ac{RSU} carries the timestamp of the most recent contributing sensor measurement, while the evaluation node records the arrival time. For each modality path (\ac{CAM}, \ac{SSL}, lidar) and the fused output, we compute the latency as the difference between this sensor timestamp and the arrival time at the evaluation node. This captures sensing, communication, and processing delays along that path, but does not assume global synchronous triggering across sensors.

In CARLA, all components share a single simulation clock, so the measured latencies primarily reflect processing time in the sensor drivers and fusion pipeline. In the \ac{CPM Lab}, each sensor runs on its own compute unit; clocks are synchronized via NTP with offsets below \qty{5}{\milli\second}, and the measured latencies additionally include realistic network and hardware effects.

Across all modalities and for the fused output, the median end-to-end latency remains below \qty{100}{\milli\second} in both CARLA and the \ac{CPM Lab}. Occasional outliers occur, for example, when multiple messages from different sources arrive in bursts, but the overall latency profile is compatible with typical requirements for cooperative driving and infrastructure-supported control. These results indicate that the same \ac{Ufil}-based pipeline can achieve comparable timing characteristics in a purely simulated environment and in a small-scale physical testbed without software-level modifications.

\begin{figure}[t]
    \centering
    \subfloat[Runtime in simulation.\label{fig:runtime_simulation}]{%
       \includegraphics{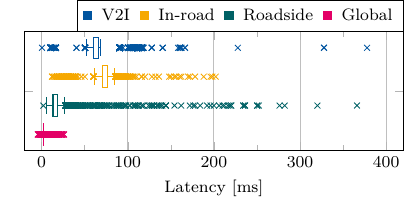}
    }
    \\
    \subfloat[Runtime in small-scale testbed.\label{fig:runtime_small_scale}]{%
        \includegraphics{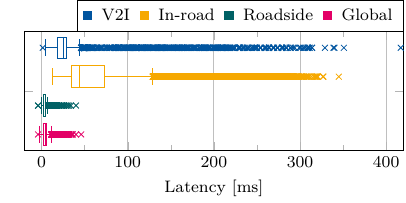}
    }
    \caption{Summary of runtime results in both simulation and small-scale testbed.}
    \label{fig:results_runtime}
\end{figure}
 
%%%%%%%%%%%%%%%%%%%%%%%%%%%%%%%%%%%%%%%%%%%%%%%%%%%%%% 
\section{Conclusion}
\label{sect:conclusion}
%%%%%%%%%%%%%%%%%%%%%%%%%%%%%%%%%%%%%%%%%%%%%%%%%%%%%% 
 
Research on infrastructure-based perception systems remains fragmented, with most existing tools tailored to ego-vehicle perception rather than static infrastructure. 
This paper presented \emph{\ac{Ufil}}, a Unified Framework for Infrastructure-based Localization, that addresses this gap by providing a standardized object model, a modular multi-object tracking pipeline, and reusable implementations of key building blocks for prediction, detection, association, update, and management.

At its core, \ac{Ufil} comprises a C++ header-only tracking library and a \ac{ROS 2} integration layer. 
The unified object representation with history buffers, together with well-defined interfaces for motion models, association metrics, solvers, and update rules, enables plug-and-play composition of tracking pipelines. 
Because all components operate on the same data model, \ac{Ufil} makes it straightforward to compare algorithms fairly, explore design trade-offs (e.g., association hypotheses vs.\ runtime), and reuse code across projects.

Using a single highway perception system as a running example, we demonstrated that the same \ac{Ufil}-based pipeline can be executed in a high-fidelity simulator and in a small-scale testbed without code changes, illustrating its scale-independent design and its suitability as a bridge between simulation and real-world deployment. Future work will extend \ac{Ufil} with additional sensing modalities and fusion strategies, setup full-scale testing scenarios, and integrate it more tightly with safety analysis and validation tools. As open-source software with public documentation and examples, \ac{Ufil} is intended as a shared research infrastructure for infrastructure-based localization rather than as a single optimized tracking algorithm.

\appendices
\balance 

\bibliographystyle{IEEEtran}
\bibliography{library.bib}

@book{aeberhard2017object,
	title        = {Object-level fusion for surround environment perception in automated driving applications},
	author       = {Aeberhard, Michael},
	year         = 2017,
	publisher    = {VDI Verlag}
}

@inproceedings{kato2018autoware,
	title        = {Autoware on board: Enabling autonomous vehicles with embedded systems},
	author       = {Kato, Shinpei and Tokunaga, Shota and Maruyama, Yuya and Maeda, Seiya and Hirabayashi, Manato and Kitsukawa, Yuki and Monrroy, Abraham and Ando, Tomohito and Fujii, Yusuke and Azumi, Takuya},
	year         = 2018,
	booktitle    = {Proceedings of the 9th ACM/IEEE International Conference on Cyber-Physical Systems (ICCPS)},
	pages        = {287--296}
}

@misc{mathworks2025automated,
	title        = {Automated Driving Toolbox},
	author       = {The MathWorks Inc.},
	year         = 2025,
	address      = {Natick, Massachusetts, United States},
	url          = {https://www.mathworks.com/help/driving/index.html},
	note         = {Version: 2025a}
}

@inproceedings{geller2024carlos,
	title        = {{CARLOS: An Open, Modular, and Scalable Simulation Framework for the Development and Testing of Software for C-ITS}},
	author       = {Geller, Christian and Haas, Benedikt and Kloeker, Amarin and Hermens, Jona and Lampe, Bastian and Beemelmanns, Till and Eckstein, Lutz},
	year         = 2024,
	booktitle    = {IEEE Intelligent Vehicles Symposium (IV)},
	pages        = {3100--3106},
	doi          = {10.1109/IV55156.2024.10588502}
}

@article{yueyuan2024choose,
	title        = {Choose Your Simulator Wisely: A Review on Open-Source Simulators for Autonomous Driving},
	author       = {Li, Yueyuan and Yuan, Wei and Zhang, Songan and Yan, Weihao and Shen, Qiyuan and Wang, Chunxiang and Yang, Ming},
	year         = 2024,
	journal      = {IEEE Transactions on Intelligent Vehicles},
	volume       = 9,
	number       = 5,
	pages        = {4861--4876},
	doi          = {10.1109/TIV.2024.3374044}
}

@inproceedings{dosovitskiy2017carla,
	title        = {{CARLA}: {An} Open Urban Driving Simulator},
	author       = {Dosovitskiy, Alexey and Ros, German and Codevilla, Felipe and Lopez, Antonio and Koltun, Vladlen},
	year         = 2017,
	month        = 11,
	booktitle    = {Proceedings of the 1st Annual Conference on Robot Learning},
	publisher    = {PMLR},
	series       = {Proceedings of Machine Learning Research},
	volume       = 78,
	pages        = {1--16},
	editor       = {Levine, Sergey and Vanhoucke, Vincent and Goldberg, Ken}
}

@inproceedings{lu2024an,
	title        = {An Extensible Framework for Open Heterogeneous Collaborative Perception},
	author       = {Lu, Yifan and Hu, Yue and Zhong, Yiqi and Wang, Dequan and Chen, Siheng and Wang, Yanfeng},
	year         = 2024,
	booktitle    = {The Twelfth International Conference on Learning Representations}
}

@article{agrawal2024semi,
	title        = {Semi-Automatic Annotation of 3D Radar and Camera for Smart Infrastructure-Based Perception},
	author       = {Agrawal, Shiva and Bhanderi, Savankumar and Elger, Gordon},
	year         = 2024,
	journal      = {IEEE Access},
	volume       = 12,
	pages        = {34325--34341},
	doi          = {10.1109/ACCESS.2024.3373310}
}

@inproceedings{vosshans2025coop,
	title        = {CoopScenes: Multi-Scene Infrastructure and Vehicle Data for Advancing Collective Perception in Autonomous Driving},
	author       = {Vosshans, Marcel and Baumann, Alexander and Drueppel, Matthias and Ait-Aider, Omar and Mezouar, Youcef and Dang, Thao and Enzweiler, Markus},
	year         = 2025,
	booktitle    = {2025 IEEE Intelligent Vehicles Symposium (IV)},
	pages        = {1040--1047},
	doi          = {10.1109/IV64158.2025.11097591}
}

@misc{polley202525dobjectdetectionintelligent,
	title        = {2.5D Object Detection for Intelligent Roadside Infrastructure},
	author       = {Nikolai Polley and Yacin Boualili and Ferdinand Mütsch and Maximilian Zipfl and Tobias Fleck and J. Marius Zöllner},
	year         = 2025,
	url          = {https://arxiv.org/abs/2507.03564},
	eprint       = {2507.03564},
	archiveprefix = {arXiv},
	primaryclass = {cs.CV}
}

@misc{karle2024edgarautonomousdrivingresearch,
	title        = {EDGAR: An Autonomous Driving Research Platform -- From Feature Development to Real-World Application},
	author       = {Phillip Karle and Tobias Betz and Marcin Bosk and Felix Fent and Nils Gehrke and Maximilian Geisslinger and Luis Gressenbuch and Philipp Hafemann and Sebastian Huber and Maximilian Hübner and Sebastian Huch and Gemb Kaljavesi and Tobias Kerbl and Dominik Kulmer and Tobias Mascetta and Sebastian Maierhofer and Florian Pfab and Filip Rezabek and Esteban Rivera and Simon Sagmeister and Leander Seidlitz and Florian Sauerbeck and Ilir Tahiraj and Rainer Trauth and Nico Uhlemann and Gerald Würsching and Baha Zarrouki and Matthias Althoff and Johannes Betz and Klaus Bengler and Georg Carle and Frank Diermeyer and Jörg Ott and Markus Lienkamp},
	year         = 2024,
	url          = {https://arxiv.org/abs/2309.15492},
	eprint       = {2309.15492},
	archiveprefix = {arXiv},
	primaryclass = {cs.RO}
}

@article{li2023set,
	title        = {Set-theoretic localization for mobile robots with infrastructure-based sensing},
	author       = {Li, Xiao and Li, Yutong and Li, Nan and Girard, Anouck and Kolmanovsky, Ilya},
	year         = 2023,
	journal      = {Advanced Control for Applications: Engineering and Industrial Systems},
	publisher    = {Wiley Online Library},
	volume       = 5,
	number       = 1,
	pages        = {e117}
}

@inproceedings{cress2022a9,
	title        = {A9-Dataset: Multi-Sensor Infrastructure-Based Dataset for Mobility Research},
	author       = {Creß, Christian and Zimmer, Walter and Strand, Leah and Fortkord, Maximilian and Dai, Siyi and Lakshminarasimhan, Venkatnarayanan and Knoll, Alois},
	year         = 2022,
	booktitle    = {2022 IEEE Intelligent Vehicles Symposium (IV)},
	volume       = {},
	number       = {},
	pages        = {965--970},
	doi          = {10.1109/IV51971.2022.9827401}
}

@inproceedings{busch2022lumpi,
	title        = {LUMPI: The Leibniz University Multi-Perspective Intersection Dataset},
	author       = {Busch, Steffen and Koetsier, Christian and Axmann, Jeldrik and Brenner, Claus},
	year         = 2022,
	booktitle    = {2022 IEEE Intelligent Vehicles Symposium (IV)},
	pages        = {1127--1134},
	organization = {IEEE}
}

@misc{urbaningv2x2025,
	title        = {UrbanIng-V2X: A Large-Scale Multi-Vehicle, Multi-Infrastructure Dataset Across Multiple Intersections for Cooperative Perception},
	author       = {Karthikeyan Chandra Sekaran and Markus Geisler and Dominik Rößle and Adithya Mohan and Daniel Cremers and Wolfgang Utschick and Michael Botsch and Werner Huber and Torsten Schön},
	year         = 2025,
	url          = {https://arxiv.org/abs/2510.23478},
	eprint       = {2510.23478},
	archiveprefix = {arXiv},
	primaryclass = {cs.CV}
}

@inproceedings{schaefer2023investigating,
	title        = {Investigating a Pressure Sensitive Surface Layer for Vehicle Localization},
	author       = {Sch\"afer, Simon and Steidl, Hendrik and Kowalewski, Stefan and Alrifaee, Bassam},
	year         = 2023,
	booktitle    = {IEEE Intelligent Vehicles Symposium (IV)},
	volume       = {},
	number       = {},
	doi          = {10.1109/IV55152.2023.10186582}
}

@misc{schaefer2025lidar,
      title={Lidar-based Tracking of Traffic Participants with Sensor Nodes in Existing Urban Infrastructure}, 
      author={Simon Schäfer and Bassam Alrifaee and Ehsan Hashemi},
      year={2025},
      eprint={2509.20009},
      archivePrefix={arXiv},
      primaryClass={cs.RO},
      url={https://arxiv.org/abs/2509.20009}, 
}

@inproceedings{schaefer2024from,
	title        = {From Small-Scale to Full-Scale: Assessing the Potential for Transferability of Experimental Results in Small-Scale CAV Testbeds},
	author       = {Sch\"afer, Simon and Alrifaee, Bassam},
	year         = 2024,
	booktitle    = {IEEE International Conference on Vehicular Electronics and Safety (ICVES)},
	volume       = {},
	number       = {},
	doi          = {10.1109/ICVES61986.2024.10927876}
}

@article{mokhtarian2024survey,
	title        = {A Survey on Small-Scale Testbeds for Connected and Automated Vehicles and Robot Swarms: A Guide for Creating a New Testbed},
	author       = {Mokhtarian, Armin and Xu, Jianye and Scheffe, Patrick and Kloock, Maximilian and Sch\"afer, Simon and Bang, Heeseung and Le, Viet-Anh and Ulhas, Sangeet and Betz, Johannes and Wilson, Sean and Berman, Spring and Paull, Liam and Prorok, Amanda and Alrifaee, Bassam},
	year         = 2024,
	journal      = {IEEE Robotics \& Automation Magazine},
	volume       = {},
	number       = {},
	doi          = {10.1109/MRA.2024.3505772}
}

@inproceedings{yu2023coca,
	title        = {Collaboration Helps Camera Overtake LiDAR in 3D Detection},
	author       = {Yue Hu and Yifan Lu and Runsheng Xu and Weidi Xie and Siheng Chen and Yanfeng Wang},
	year         = 2023,
	booktitle    = {The IEEE/CVF Conference on Computer Vision and Pattern Recognition (CVPR)}
}

@inproceedings{schubert2008comparison,
	title        = {Comparison and evaluation of advanced motion models for vehicle tracking},
	author       = {Schubert, Robin and Richter, Eric and Wanielik, Gerd},
	year         = 2008,
	booktitle    = {2008 11th International Conference on Information Fusion},
	volume       = {},
	number       = {},
	pages        = {1--6},
	doi          = {}
}

@book{barshalom2002estimation,
	title        = {Estimation with applications to tracking and navigation: theory algorithms and software},
	author       = {Bar-Shalom, Yaakov and Li, X Rong and Kirubarajan, Thiagalingam},
	year         = 2002,
	publisher    = {John Wiley \& Sons},
	doi          = {10.1002/0471221279}
}

@book{bucy1987filtering,
	title        = {Filtering for stochastic processes with applications to guidance},
	author       = {Bucy, Richard S. and Joseph, Peter D.},
	year         = 1987,
	publisher    = {Chelsea Pub. Co},
	place        = {New York, NY}
}

@inproceedings{kueppers2024v2aix,
	title        = {V2AIX: A Multi-Modal Real-World Dataset of ETSI ITS V2X Messages in Public Road Traffic},
	author       = {Kueppers, Guido and Busch, Jean-Pierre and Reiher, Lennart and Eckstein, Lutz},
	year         = 2024,
	booktitle    = {IEEE International Conference on Intelligent Transportation Systems (ITSC)},
	volume       = {},
	number       = {},
	pages        = {392--398},
	doi          = {10.1109/ITSC58415.2024.10920150}
}

@inproceedings{kloock2021cyberphysical,
	title        = {Cyber-Physical Mobility Lab: An Open-Source Platform for Networked and Autonomous Vehicles},
	author       = {Kloock, Maximilian and Scheffe, Patrick and Maczijewski, Janis and Kampmann, Alexandru and Mokhtarian, Armin and Kowalewski, Stefan and Alrifaee, Bassam},
	year         = 2021,
	booktitle    = {EUCA European Control Conference (ECC)},
	volume       = {},
	number       = {},
	pages        = {1937--1944},
	doi          = {10.23919/ECC54610.2021.9654986}
}

@article{scheffe2020networked,
	title        = {Networked and Autonomous Model-scale Vehicles for Experiments in Research and Education},
	author       = {Patrick Scheffe and Janis Maczijewski and Maximilian Kloock and Alexandru Kampmann and Andreas Derks and Stefan Kowalewski and Bassam Alrifaee},
	year         = 2020,
	journal      = {IFAC-PapersOnLine},
	publisher    = {Elsevier},
	volume       = 53,
	number       = 2,
	pages        = {17332--17337},
	doi          = {10.1016/j.ifacol.2020.12.1821}
}

@inproceedings{haibao2022dairv2x,
	title        = {Dair-v2x: A large-scale dataset for vehicle-infrastructure cooperative 3d object detection},
	author       = {Yu, Haibao and Luo, Yizhen and Shu, Mao and Huo, Yiyi and Yang, Zebang and Shi, Yifeng and Guo, Zhenglong and Li, Hanyu and Hu, Xing and Yuan, Jirui and Nie, Zaiqing},
	year         = 2022,
	booktitle    = {Proceedings of the IEEE/CVF Conference on Computer Vision and Pattern Recognition},
	pages        = {21361--21370}
}

@article{kloock2020vision,
	title        = {Vision-based real-time indoor positioning system for multiple vehicles},
	author       = {Kloock, Maximilian and Scheffe, Patrick and T{\"u}lleners, Isabelle and Maczijewski, Janis and Kowalewski, Stefan and Alrifaee, Bassam},
	year         = 2020,
	journal      = {IFAC-PapersOnLine},
	publisher    = {Elsevier},
	volume       = 53,
	number       = 2,
	pages        = {15446--15453},
	doi          = {10.1016/j.ifacol.2020.12.2367}
}

\end{document}